\documentstyle[colap,ifthen,calc]{article}

\begin{document}

 
\def\diatop[#1|#2]{{\setbox1=\hbox{{#1{}}}\setbox2=\hbox{{#2{}}}%
                    \dimen0=\ifdim\wd1>\wd2\wd1\else\wd2\fi%
                    \dimen1=\ht2\advance\dimen1by-1ex%
                    \setbox1=\hbox to1\dimen0{\hss#1\hss}%
                    \rlap{\raise1\dimen1\box1}%
                    \hbox to1\dimen0{\hss#2\hss}}}%
 
 
 
\font\ipatwelverm=wsuipa12
\def\ipa{\ipatwelverm}
 
\def\inva{\edef\next{\the\font}\ipa\char'000\next}%
\def\scripta{\edef\next{\the\font}\ipa\char'001\next}%
\def\nialpha{\edef\next{\the\font}\ipa\char'002\next}%
\def\invscripta{\edef\next{\the\font}\ipa\char'003\next}%
\def\invv{\edef\next{\the\font}\ipa\char'004\next}%
 
\def\crossb{\edef\next{\the\font}\ipa\char'005\next}%
\def\barb{\edef\next{\the\font}\ipa\char'006\next}%
\def\slashb{\edef\next{\the\font}\ipa\char'007\next}%
\def\hookb{\edef\next{\the\font}\ipa\char'010\next}%
\def\nibeta{\edef\next{\the\font}\ipa\char'011\next}%
 
\def\slashc{\edef\next{\the\font}\ipa\char'012\next}%
\def\curlyc{\edef\next{\the\font}\ipa\char'013\next}%
\def\clickc{\edef\next{\the\font}\ipa\char'014\next}%
 
\def\crossd{\edef\next{\the\font}\ipa\char'015\next}%
\def\bard{\edef\next{\the\font}\ipa\char'016\next}%
\def\slashd{\edef\next{\the\font}\ipa\char'017\next}%
\def\hookd{\edef\next{\the\font}\ipa\char'020\next}%
\def\taild{\edef\next{\the\font}\ipa\char'021\next}%
\def\dz{\edef\next{\the\font}\ipa\char'022\next}%
\def\eth{\edef\next{\the\font}\ipa\char'023\next}%
\def\scd{\edef\next{\the\font}\ipa\char'024\next}%
 
\def\schwa{\edef\next{\the\font}\ipa\char'025\next}%
\def\er{\edef\next{\the\font}\ipa\char'026\next}%
\def\reve{\edef\next{\the\font}\ipa\char'027\next}%
\def\niepsilon{\edef\next{\the\font}\ipa\char'030\next}%
\def\revepsilon{\edef\next{\the\font}\ipa\char'031\next}%
\def\hookrevepsilon{\edef\next{\the\font}\ipa\char'032\next}%
\def\closedrevepsilon{\edef\next{\the\font}\ipa\char'033\next}%
 
\def\scriptg{\edef\next{\the\font}\ipa\char'034\next}%
\def\hookg{\edef\next{\the\font}\ipa\char'035\next}%
\def\scg{\edef\next{\the\font}\ipa\char'036\next}%
\def\nigamma{\edef\next{\the\font}\ipa\char'037\next}
\def\ipagamma{\edef\next{\the\font}\ipa\char'040\next}%
\def\babygamma{\edef\next{\the\font}\ipa\char'041\next}%
 
\def\hv{\edef\next{\the\font}\ipa\char'042\next}%
\def\crossh{\edef\next{\the\font}\ipa\char'043\next}%
\def\hookh{\edef\next{\the\font}\ipa\char'044\next}%
\def\hookheng{\edef\next{\the\font}\ipa\char'045\next}%
\def\invh{\edef\next{\the\font}\ipa\char'046\next}%
 
\def\bari{\edef\next{\the\font}\ipa\char'047\next}%
\def\dlbari{\edef\next{\the\font}\ipa\char'050\next}
\def\niiota{\edef\next{\the\font}\ipa\char'051\next}%
\def\sci{\edef\next{\the\font}\ipa\char'052\next}%
\def\barsci{\edef\next{\the\font}\ipa\char'053\next}
 
\def\invf{\edef\next{\the\font}\ipa\char'054\next}%
 
\def\tildel{\edef\next{\the\font}\ipa\char'055\next}%
\def\barl{\edef\next{\the\font}\ipa\char'056\next}%
\def\latfric{\edef\next{\the\font}\ipa\char'057\next}%
\def\taill{\edef\next{\the\font}\ipa\char'060\next}%
\def\lz{\edef\next{\the\font}\ipa\char'061\next}%
\def\nilambda{\edef\next{\the\font}\ipa\char'062\next}%
\def\crossnilambda{\edef\next{\the\font}\ipa\char'063\next}%
 
\def\labdentalnas{\edef\next{\the\font}\ipa\char'064\next}%
\def\invm{\edef\next{\the\font}\ipa\char'065\next}%
\def\legm{\edef\next{\the\font}\ipa\char'066\next}%
 
\def\nj{\edef\next{\the\font}\ipa\char'067\next}%
\def\eng{\edef\next{\the\font}\ipa\char'070\next}%
\def\tailn{\edef\next{\the\font}\ipa\char'071\next}%
\def\scn{\edef\next{\the\font}\ipa\char'072\next}%
 
\def\clickb{\edef\next{\the\font}\ipa\char'073\next}%
\def\baro{\edef\next{\the\font}\ipa\char'074\next}%
\def\openo{\edef\next{\the\font}\ipa\char'075\next}%
\def\niomega{\edef\next{\the\font}\ipa\char'076\next}%
\def\closedniomega{\edef\next{\the\font}\ipa\char'077\next}%
\def\oo{\edef\next{\the\font}\ipa\char'100\next}%
 
\def\barp{\edef\next{\the\font}\ipa\char'101\next}%
\def\thorn{\edef\next{\the\font}\ipa\char'102\next}%
\def\niphi{\edef\next{\the\font}\ipa\char'103\next}%
 
\def\flapr{\edef\next{\the\font}\ipa\char'104\next}%
\def\legr{\edef\next{\the\font}\ipa\char'105\next}%
\def\tailr{\edef\next{\the\font}\ipa\char'106\next}%
\def\invr{\edef\next{\the\font}\ipa\char'107\next}%
\def\tailinvr{\edef\next{\the\font}\ipa\char'110\next}%
\def\invlegr{\edef\next{\the\font}\ipa\char'111\next}%
\def\scr{\edef\next{\the\font}\ipa\char'112\next}%
\def\invscr{\edef\next{\the\font}\ipa\char'113\next}%
 
\def\tails{\edef\next{\the\font}\ipa\char'114\next}%
\def\esh{\edef\next{\the\font}\ipa\char'115\next}%
\def\curlyesh{\edef\next{\the\font}\ipa\char'116\next}%
\def\nisigma{\edef\next{\the\font}\ipa\char'117\next}%
 
\def\tailt{\edef\next{\the\font}\ipa\char'120\next}%
\def\tesh{\edef\next{\the\font}\ipa\char'121\next}%
\def\clickt{\edef\next{\the\font}\ipa\char'122\next}%
\def\nitheta{\edef\next{\the\font}\ipa\char'123\next}%
 
\def\baru{\edef\next{\the\font}\ipa\char'124\next}%
\def\slashu{\edef\next{\the\font}\ipa\char'125\next}%
\def\niupsilon{\edef\next{\the\font}\ipa\char'126\next}%
\def\scu{\edef\next{\the\font}\ipa\char'127\next}%
\def\barscu{\edef\next{\the\font}\ipa\char'130\next}%
 
\def\scriptv{\edef\next{\the\font}\ipa\char'131\next}%
 
\def\invw{\edef\next{\the\font}\ipa\char'132\next}%
 
\def\nichi{\edef\next{\the\font}\ipa\char'133\next}%
 
\def\invy{\edef\next{\the\font}\ipa\char'134\next}%
\def\scy{\edef\next{\the\font}\ipa\char'135\next}%
 
\def\curlyz{\edef\next{\the\font}\ipa\char'136\next}%
\def\tailz{\edef\next{\the\font}\ipa\char'137\next}%
\def\yogh{\edef\next{\the\font}\ipa\char'140\next}%
\def\curlyyogh{\edef\next{\the\font}\ipa\char'141\next}%
 
\def\glotstop{\edef\next{\the\font}\ipa\char'142\next}%
\def\revglotstop{\edef\next{\the\font}\ipa\char'143\next}%
\def\invglotstop{\edef\next{\the\font}\ipa\char'144\next}%
\def\ejective{\edef\next{\the\font}\ipa\char'145\next}%
\def\reveject{\edef\next{\the\font}\ipa\char'146\next}%
 
 
\def\dental#1{\oalign{#1\crcr
          \hidewidth{\ipa\char'147}\hidewidth}}
 
\def\upt{\edef\next{\the\font}\ipa\char'154\next}
\def\downt{\edef\next{\the\font}\ipa\char'155\next}%
\def\leftt{\edef\next{\the\font}\ipa\char'156\next}%
\def\rightt{\edef\next{\the\font}\ipa\char'157\next}%
 
\def\upp{\edef\next{\the\font}\ipa\char'164\next}
\def\downp{\edef\next{\the\font}\ipa\char'165\next}%
\def\leftp{\edef\next{\the\font}\ipa\char'166\next}%
\def\rightp{\edef\next{\the\font}\ipa\char'167\next}%
 
\def\stress{\edef\next{\the\font}\ipa\char'150\next}
\def\secstress{\edef\next{\the\font}\ipa\char'151\next}
 
\def\syllabic{\edef\next{\the\font}\ipa\char'152\next}
 
\def\corner{\edef\next{\the\font}\ipa\char'153\next}%
 
\def\halflength{\edef\next{\the\font}\ipa\char'160\next}
\def\length{\edef\next{\the\font}\ipa\char'161\next}
 
\def\underdots{\edef\next{\the\font}\ipa\char'162\next}%
 
\def\ain{\edef\next{\the\font}\ipa\char'163\next}
 
\def\overring{\edef\next{\the\font}\ipa\char'170\next}%
\def\underring{\edef\next{\the\font}\ipa\char'171\next}%
 
\def\open{\edef\next{\the\font}\ipa\char'172\next}%
 
\def\midtilde{\edef\next{\the\font}\ipa\char'173\next}%
\def\undertilde{\edef\next{\the\font}\ipa\char'174\next}%
 
\def\underwedge{\edef\next{\the\font}\ipa\char'175\next}%
 
\def\polishhook{\edef\next{\the\font}\ipa\char'176\next}%
 
\def\underarch#1{\oalign{#1\crcr
          \hidewidth{\ipa\char'177}\hidewidth}}
 

\font\ipatenrm=wsuipa10
\def\ipa{\ipatenrm}

\newcommand{\A}{\ejective}  
\newcommand{\h}{\crossh}    
\newcommand{\CC}{\reveject} 
\newcommand{\sh}{\v{s}}	    
\newcommand{\J}{\^{\j}}     
\newcommand{\TT}{\d{t}}   
\newcommand{\Ss}{\d{s}}   
\newcommand{\OO}{\^{o}}      

\newcommand{\br}{\b{b}}   
\newcommand{\gr}{\b{g}}
\newcommand{\dr}{\b{d}}
\newcommand{\kr}{\b{k}}
\newcommand{\pr}{\b{p}}
\newcommand{\tr}{\b{t}}
\newcommand{\e}{\schwa}


\newcommand{\Ru}{{\em Rukk\={a}\kr\^{a}}}  
\newcommand{\Qu}{{\em Qu\sh\sh\={a}y\^{a}}}  
\newcommand{\bgdkpt}{{\em b\gr\={a}\dr k\pr\={a}\tr}}


\newcommand{\Syl}{$\sigma$}             
\newcommand{\Mor}{$\mu$}                
\newcommand{\Sylm}{$\sigma_{\mu}$}      
\newcommand{\Sylmm}{$\sigma_{\mu\mu}$}  
\newcommand{\Sylx}{$\sigma_{x}$} 
\newcommand{\kernel}{B:$\Phi$}          
\newcommand{\residue}{B/$\Phi$}         
\newcommand{\ppc}{O:$\Phi$}             
\newcommand{\npc}{O/$\Phi$}             

\newcommand{\composition}{$\circ$}      
\newcommand{\conc}{$^{\frown}$}         
\newcommand{\estr}{$\varepsilon$}   
\newcommand{\func}[2]
   {\mbox{{\sc #1(}#2{\sc )}}}

\newcommand{\lab}{$\langle$} 
\newcommand{\rab}{$\rangle$} 


\newcounter{boxwidth}
\newcounter{boxhight}
\newcounter{mttlmboxwidth}
\newcounter{mttlmboxhight}
\newcounter{autosegboxwidth}
\newcounter{defaulthight}
\newcounter{strlength}
\newcounter{notiers}
\newcounter{picwidth}
\newcounter{pichight}
\newcounter{lift}
\newcounter{linelen}
\newcounter{xdirection}
\newcounter{ydirection}
\newcounter{curx}
\newcounter{cury}
\newcounter{pictopmargin}

\newlength{\tapenamewidth}
\newlength{\templen}

\setcounter{mttlmboxwidth}{12}%
\setcounter{mttlmboxhight}{12}
\setcounter{autosegboxwidth}{10}%
\setcounter{pictopmargin}{2}

\newcounter{fsmboxwidth}  \setcounter{fsmboxwidth}{30}
\newcounter{fsmcolumns}  \setcounter{fsmcolumns}{7}
\newcounter{fsmrows}
\newcounter{fsmcolumnsx}
\newcounter{fsmrowsx}
\newcounter{nostates}
\newcounter{curstate}
\newcounter{fsmradius}

\newcommand{\fsm}[2]
   {
    \setcounter{boxwidth}{\value{fsmboxwidth}}%
    \setcounter{boxhight}{\value{fsmboxwidth}}%
    \setcounter{nostates}{0}%
    \countstates#1|| END%
    \setcounter{fsmrows}{\value{nostates}/\value{fsmcolumns}}%
    \setcounter{temp}{\value{fsmrows}*\value{fsmcolumns}}%
    \ifthenelse{\value{nostates} = \value{temp}}%
               {}%
               {\stepcounter{fsmrows}}%
    \setcounter{fsmcolumnsx}{\value{fsmcolumns}-1}%
    \setcounter{fsmrowsx}{\value{fsmrows}-1}%
    \setcounter{fsmradius}{\value{boxwidth}-\value{boxwidth}/10}%
    \setcounter{picwidth}{\value{boxwidth}*\value{fsmcolumns}+%
                          \value{boxwidth}*\value{fsmcolumnsx}+%
                          \value{boxwidth}*2}%
    \setcounter{pichight}{\value{boxhight}*\value{fsmrows}+%
                          \value{boxhight}*\value{fsmrowsx}+%
                          \value{boxhight}*2}%
    \setcounter{lift}{\value{pichight}/-2}%
    \rule[\the\value{lift}pt]{3 pt}{\the\value{pichight}pt}%
    \begin{picture}(\the\value{picwidth},0)(0,-\the\value{lift})%
       \setcounter{curstate}{0}%
       \setcounter{curx}{\value{boxhight}}%
       \setcounter{cury}{\value{boxhight}*\value{fsmrowsx}*2+\value{boxhight}}%
       \drawstates#1|| END%
    \end{picture}
   }

\def\drawstates#1,#2,#3|#4 END
   {
    \put(\the\value{curx},\the\value{cury})%
         {\makebox(\the\value{boxwidth},\the\value{boxhight}){#1}}%
    \setcounter{curx}{\value{curx}+\value{boxwidth}/2}%
    \setcounter{cury}{\value{cury}+\value{boxhight}/2}%
    \put(\the\value{curx},\the\value{cury})%
         {\circle{\the\value{boxwidth}}}%
    \ifthenelse{\equal{#3}{y}}%
       {\put(\the\value{curx},\the\value{cury})%
            {\circle{\the\value{fsmradius}}}}%
       {}%
    \setcounter{curx}{\value{curx}-\value{boxwidth}/2}%
    \setcounter{cury}{\value{cury}-\value{boxhight}/2}%
    \stepcounter{curstate}%
    \ifthenelse{\value{curstate} = \value{fsmcolumns}}
               {\setcounter{curx}{\value{boxhight}}%
                \setcounter{cury}{\value{cury}-\value{boxhight}*2}%
                \setcounter{curstate}{0}}%
               {\setcounter{curx}{\value{curx}+\value{boxwidth}*2}}%
    \ifthenelse{\equal{#4}{|}}{}%
               {\drawstates#4 END}%
   }

\def\countstates#1|#2 END
   {\stepcounter{nostates}%
    \ifthenelse{\equal{#2}{|}}{}%
               {\countstates#2 END}%
   }


\newcommand{\mttlmsetwidth}[1]%
   {\setcounter{mttlmboxwidth}{#1}}

\newcommand{\mttlmsethight}[1]%
   {\setcounter{mttlmboxhight}{#1}}

\newcommand{\mttlm}[3]
   {\immediate\write16{MTTLM = #1}%
    \setcounter{boxwidth}{\value{mttlmboxwidth}}%
    \setcounter{boxhight}{\value{mttlmboxhight}}%
    \setcounter{strlength}{0}%
    \setcounter{notiers}{0}%
    \setlength{\tapenamewidth}{0pt}%
    \counttiers#3|| END%
    \countsurfacewidth#1 END%
    \findtapenamewidth#3|| END%
    \setcounter{picwidth}{\value{boxwidth}*\value{strlength}}%
    \setcounter{pichight}{\value{boxhight}*\value{notiers}+%
                          2*\value{boxhight}+\value{pictopmargin}}%
    \setcounter{lift}{\value{pichight}/-2}%
    \rule[\the\value{lift}pt]{0 pt}{\the\value{pichight}pt}%
    \begin{picture}(\the\value{picwidth},0)(0,-\the\value{lift})%
       \let\boxtype=\makebox%
       \setcounter{curx}{0}%
       \setcounter{cury}{0}%
       \displayonetape#1 END%
       \setcounter{curx}{0}%
       \setcounter{cury}{\value{boxhight}}%
       \displaystrings#2|| END%
       \setcounter{curx}{0}%
       \setcounter{cury}{2*\value{boxhight}}%
       \displaymanytapes#3|| END%
    \end{picture}%
    \hspace{\the\tapenamewidth}%
   }

\def\countsurfacewidth#1:#2 END%
   {\countwidth#1| END%
    \maxtapenamewidth#2 END%
   }

\def\displayonetape#1:#2 END%
   {
    \put(\the\value{curx},\the\value{cury})%
         {\framebox(\the\value{picwidth},\the\value{boxhight}){}}%
    \partition#1| END%
    \put(\the\value{curx},\the\value{cury})%
         {\makebox(\the\value{boxwidth},\the\value{boxhight})[l]{\ {\em #2}}}%
    \setcounter{curx}{0}%
    \displaystrings#1|| END%
   }    

\def\displaymanytapes#1|#2 END%
   {\displayonetape#1 END%
    \addtocounter{cury}{-\value{boxhight}}%
    \ifthenelse{\equal{#2}{|}}{}%
               {\displaymanytapes#2 END}%
   }

\def\partition#1#2 END%
   {\addtocounter{curx}{\value{boxwidth}}%
    \ifthenelse{\equal{#1}{-}}{}%
                      {\put(\value{curx},\value{cury})%
                         {\dashbox{0.5}(0,\value{boxhight}){}}}
    \ifthenelse{\equal{#2}{|}}{}%
               {\partition#2 END}%
   }

\def\findtapenamewidth#1:#2|#3 END
   {\maxtapenamewidth#2 END%
    \ifthenelse{\equal{#3}{|}}{}%
               {\findtapenamewidth#3 END}%
   }

\def\maxtapenamewidth#1 END%
   {\settowidth{\templen}{{\em #1}}%
    \ifthenelse{\templen > \tapenamewidth}%
               {\settowidth{\tapenamewidth}{\ {\em #1}}}%
               {}%
   }


\newcommand{\autosegsetwidth}[1]%
   {\setcounter{autosegboxwidth}{#1}}

\newcommand{\autoseg}[2]%
   {\immediate\write16{Autseg Tiers #1}%
    \setcounter{boxwidth}{\value{autosegboxwidth}}%
    \setcounter{boxhight}{\value{autosegboxwidth}}%
    \setcounter{strlength}{0}%
    \setcounter{notiers}{0}%
    \counttiers#1|| END%
    \countwidth#1 END%
    \setcounter{picwidth}{\value{boxwidth}*\value{strlength}}%
    \setcounter{pichight}{2*\value{boxwidth}*\value{notiers}-%
                          \value{boxwidth}+\value{pictopmargin}}%
    \setcounter{lift}{\value{pichight}/-2}%
    \rule[\the\value{lift}pt]{0 pt}{\the\value{pichight}pt}%
    \begin{picture}(\the\value{picwidth},0)(0,-\the\value{lift})%
       \let\boxtype=\makebox%
       \setcounter{curx}{0}%
       \setcounter{cury}{0}%
       \displaystrings#1|| END%
       \setcounter{curx}{\value{boxwidth}/2}%
       \linkstrings#2,||| END%
    \end{picture}%
   }

\def\counttiers#1|#2 END
   {\stepcounter{notiers}%
    \ifthenelse{\equal{#2}{|}}{}%
               {\counttiers#2 END}%
   }

\def\countwidth#1|#2 END%
   {\countlength#1| END%
   }

\def\countlength#1#2 END
   {\stepcounter{strlength}%
    \ifthenelse{\equal{#2}{|}}{}%
               {\countlength#2 END}%
   }

\def\displaystrings#1|#2 END
   {\setmorpheme#1| END%
    \setcounter{curx}{0}%
    \addtocounter{cury}{2*\value{boxhight}}%
    \ifthenelse{\equal{#2}{|}}{}%
               {\displaystrings#2 END}%
   }

\def\setmorpheme#1#2 END%
   {\ifthenelse{\equal{#1}{-}}{}%
               {\put(\the\value{curx},\the\value{cury})%
                {\boxtype(\the\value{boxwidth},\the\value{boxhight}){#1}}}%
    \addtocounter{curx}{\value{boxwidth}}%
    \ifthenelse{\equal{#2}{|}}{}%
               {\setmorpheme#2 END}%
   }

\def\linkstrings#1#2#3,#4 END
   {
    \ifthenelse{\equal{#1}{-}}{}%
       {
        \ifthenelse{#1 < #2}%
               {\setcounter{cury}{(2*#1-1)*\value{boxwidth}}%
                \setcounter{ydirection}{1}%
                \setcounter{linelen}{(2*(#2-#1)-1)*\value{boxwidth}}}%
               {}%
        \ifthenelse{#1 > #2}%
               {\setcounter{cury}{2*(#1-1)*\value{boxwidth}}%
                \setcounter{ydirection}{-1}%
                \setcounter{linelen}{(2*(#1-#2)-1)*\value{boxwidth}}}%
               {}%
        \setcounter{xdirection}{0}%
        \ifthenelse{\equal{#3}{r}}{\setcounter{xdirection}{1}}{}%
        \ifthenelse{\equal{#3}{rr}}{\setcounter{xdirection}{2}%
                                   \addtocounter{linelen}{\value{boxwidth}}}{}%
        \ifthenelse{\equal{#3}{rrr}}{\setcounter{xdirection}{3}%
                                   \addtocounter{linelen}{2*\value{boxwidth}}}%
                                   {}%
        \ifthenelse{\equal{#3}{l}}{\setcounter{xdirection}{-1}}{}%
        \ifthenelse{\equal{#3}{ll}}{\setcounter{xdirection}{-2}%
                                   \addtocounter{linelen}{\value{boxwidth}}}{}%
        \ifthenelse{\equal{#3}{lll}}{\setcounter{xdirection}{-3}%
                                   \addtocounter{linelen}{2*\value{boxwidth}}}%
                                   {}%
        \put(\value{curx},\value{cury})%
            {\line(\value{xdirection},\value{ydirection})%
            {\value{linelen}}}}%
    \addtocounter{curx}{\value{boxwidth}}%
    \ifthenelse{\equal{#4}{|||}}{}%
               {\linkstrings#4 END}%
   }

\newcounter{moraboxwidth}
\setcounter{moraboxwidth}{10}

\newcommand{\morasetwidth}[1]%
   {\setcounter{moraboxwidth}{#1}}

\newcommand{\moratree}[2]%
   {
    \setcounter{picwidth}{2*\value{moraboxwidth}}%
    \setcounter{pichight}{4*\value{moraboxwidth}+\value{moraboxwidth}/2}%
    \setcounter{lift}{\value{pichight}/-2}%
    \rule[\the\value{lift}pt]{0 pt}{\the\value{pichight}pt}%
    \begin{picture}(\the\value{picwidth},0)(0,-\the\value{lift})%
       \drawbasic{#1}{#2}%
    \end{picture}%
   }

\newcommand{\mmoratree}[3]%
   {
    \setcounter{picwidth}{3*\value{moraboxwidth}}%
    \setcounter{pichight}{4*\value{moraboxwidth}+\value{moraboxwidth}/2}%
    \setcounter{lift}{\value{pichight}/-2}%
    \rule[\the\value{lift}pt]{0 pt}{\the\value{pichight}pt}%
    \begin{picture}(\the\value{picwidth},0)(0,-\the\value{lift})%
       \drawbasic{#1}{#2}%
       \setcounter{curx}{2*\value{moraboxwidth}}%
       \put(\value{curx},0)%
          {\makebox(\value{moraboxwidth},\value{moraboxwidth})[b]{#3}}%
       \setcounter{cury}{2*\value{moraboxwidth}}%
       \put(\value{curx},\value{cury})%
          {\makebox(\value{moraboxwidth},\value{moraboxwidth}){\Mor}}%
       \setcounter{curx}{2*\value{moraboxwidth}+\value{moraboxwidth}/2}%
       \put(\value{curx},\value{cury}){\line(0,-1){\value{moraboxwidth}}}%
       \setcounter{curx}{\value{moraboxwidth}+\value{moraboxwidth}/2}%
       \setcounter{cury}{3*\value{moraboxwidth}+\value{moraboxwidth}/2}%
       \put(\value{curx},\value{cury}){\line(2,-1){\value{moraboxwidth}}}%
    \end{picture}%
   }

\newcommand{\xmoratree}[1]%
   {
    \setcounter{picwidth}{\value{moraboxwidth}}%
    \setcounter{pichight}{4*\value{moraboxwidth}+\value{moraboxwidth}/2}%
    \setcounter{lift}{\value{pichight}/-2}%
    \rule[\the\value{lift}pt]{0 pt}{\the\value{pichight}pt}%
    \begin{picture}(\the\value{picwidth},0)(0,-\the\value{lift})%
       \put(0,0){\makebox(\value{moraboxwidth},\value{moraboxwidth})[b]{#1}}%
       \setcounter{cury}{3*\value{moraboxwidth}+\value{moraboxwidth}/2}%
       \put(0,\value{cury})%
       {\makebox(\value{moraboxwidth},\value{moraboxwidth}){\Sylx}}%
       \setcounter{curx}{\value{moraboxwidth}/2}%
       \setcounter{linelen}{2*\value{moraboxwidth}+\value{moraboxwidth}/2}%
       \put(\value{curx},\value{cury}){\line(0,-1){\value{linelen}}}%
    \end{picture}%
   }

\newcommand{\gmoratree}[5]%
   {\mbox{\mmoratree{#1}{#2}{#3}%
          \hspace{-\value{moraboxwidth}pt}%
          \mmoratree{{}}{#4}{#5}}%
   \immediate\write16{(#1,#2,#3,#4,#5)}%
   }

\newcommand{\drawbasic}[2]%
   {
    \put(0,0){\makebox(\value{moraboxwidth},\value{moraboxwidth})[b]{#1}}%
    \put(\value{moraboxwidth},0)%
       {\makebox(\value{moraboxwidth},\value{moraboxwidth})[b]{#2}}%
    \setcounter{cury}{2*\value{moraboxwidth}}%
    \put(\value{moraboxwidth},\value{cury})%
       {\makebox(\value{moraboxwidth},\value{moraboxwidth}){\Mor}}%
    \setcounter{cury}{3*\value{moraboxwidth}+\value{moraboxwidth}/2}%
    \put(\value{moraboxwidth},\value{cury})%
       {\makebox(\value{moraboxwidth},\value{moraboxwidth}){\Syl}}%
    \setcounter{curx}{\value{moraboxwidth}+\value{moraboxwidth}/2}%
    \setcounter{cury}{3*\value{moraboxwidth}+\value{moraboxwidth}/2}%
    \put(\value{curx},\value{cury}){\line(-2,-5){\value{moraboxwidth}}}%
    \setcounter{linelen}{\value{moraboxwidth}/2}%
    \put(\value{curx},\value{cury}){\line(0,-1){\value{linelen}}}%
    \setcounter{cury}{2*\value{moraboxwidth}}%
    \put(\value{curx},\value{cury}){\line(0,-1){\value{moraboxwidth}}}%
   }

\newcounter{tapeboxhight}
\setcounter{tapeboxhight}{15}
\newcounter{delta}
\newcounter{fstwidth}
\newcounter{temp}

\newcommand{\tapehight}[1]%
   {\setcounter{tapeboxhight}{#1}}

\newcommand{\cascadetransducers}%
   {
    \setcounter{picwidth}{10*\value{tapeboxhight}}%
    \setcounter{pichight}{9*\value{tapeboxhight}}%
    \setcounter{lift}{\value{pichight}/-2}%
    \rule[\the\value{lift}pt]{0 pt}{\the\value{pichight}pt}%
    \begin{picture}(\the\value{picwidth},0)(0,-\the\value{lift})%
       \put(0,0){\framebox(\value{picwidth},\value{tapeboxhight})
             {Surface String}}%
       \setcounter{cury}{4*\value{tapeboxhight}}
       \setcounter{temp}{8*\value{tapeboxhight}}
       \put(\value{tapeboxhight},\value{cury})
             {\framebox(\value{temp},\value{tapeboxhight})
             {Intermediate String}}%
       \setcounter{cury}{8*\value{tapeboxhight}}
       \put(0,\value{cury}){\framebox(\value{picwidth},\value{tapeboxhight})
             {Lexical String}}%
       \setcounter{curx}{\value{picwidth}/2}
       \setcounter{cury}{\value{tapeboxhight}}
       \setcounter{delta}{2*\value{tapeboxhight}}
       \multiput(\value{curx},\value{cury})(0,\value{delta}){4}
          {\line(0,1){\value{tapeboxhight}}}
       \setcounter{cury}{2*\value{tapeboxhight}+\value{tapeboxhight}/2}
       \setcounter{delta}{4*\value{tapeboxhight}}
       \setcounter{fstwidth}{2*\value{tapeboxhight}}
       \multiput(\value{curx},\value{cury})(0,\value{delta}){2}
          {\oval(\value{fstwidth},\value{tapeboxhight})}
       \put(\value{curx},\value{cury}){\makebox(0,0){\em FST$_n$}}%
       \setcounter{cury}{6*\value{tapeboxhight}+\value{tapeboxhight}/2}
       \put(\value{curx},\value{cury}){\makebox(0,0){\em FST$_1$}}%
    \end{picture}%
    ~$\Longrightarrow$~%
    \begin{picture}(\the\value{picwidth},0)(0,-\the\value{lift})%
       \put(0,0){\framebox(\value{picwidth},\value{tapeboxhight})
             {Surface String}}%
       \setcounter{cury}{8*\value{tapeboxhight}}
       \put(0,\value{cury}){\framebox(\value{picwidth},\value{tapeboxhight})
             {Lexical String}}%
       \setcounter{curx}{\value{picwidth}/2}
       \setcounter{cury}{\value{tapeboxhight}}
       \setcounter{temp}{3*\value{tapeboxhight}}
       \setcounter{delta}{4*\value{tapeboxhight}}
       \multiput(\value{curx},\value{cury})(0,\value{delta}){2}
          {\line(0,1){\value{temp}}}
       \setcounter{cury}{4*\value{tapeboxhight}+\value{tapeboxhight}/2}
       \setcounter{fstwidth}{8*\value{tapeboxhight}}
       \put(\value{curx},\value{cury})
              {\oval(\value{fstwidth},\value{tapeboxhight})}
       \put(\value{curx},\value{cury})
              {\makebox(0,0){$FST_1 \circ FST_2 \circ \cdots \circ FST_n$}}%
    \end{picture}%
   }

\newcommand{\paralleltransducers}%
   {\paralleltransducersone%
    ~$\Longrightarrow$~%
    \paralleltransducerstwo%
   }

\newcommand{\paralleltransducersone}%
   {
    \setcounter{picwidth}{10*\value{tapeboxhight}}%
    \setcounter{pichight}{9*\value{tapeboxhight}}%
    \setcounter{lift}{\value{pichight}/-2}%
    \rule[\the\value{lift}pt]{0 pt}{\the\value{pichight}pt}%
    \begin{picture}(\the\value{picwidth},0)(0,-\the\value{lift})%
       \put(0,0){\framebox(\value{picwidth},\value{tapeboxhight})
             {Surface String}}%
       \setcounter{cury}{8*\value{tapeboxhight}}
       \put(0,\value{cury}){\framebox(\value{picwidth},\value{tapeboxhight})
             {Lexical String}}%
       \setcounter{curx}{\value{picwidth}/2}
       \setcounter{cury}{\value{tapeboxhight}}
       \setcounter{delta}{6*\value{tapeboxhight}}
       \multiput(\value{curx},\value{cury})(0,\value{delta}){2}
          {\line(0,1){\value{tapeboxhight}}}
       \setcounter{curx}{\value{tapeboxhight}}
       \setcounter{cury}{2*\value{tapeboxhight}}
       \setcounter{delta}{5*\value{tapeboxhight}}
       \setcounter{temp}{8*\value{tapeboxhight}}
       \multiput(\value{curx},\value{cury})(0,\value{delta}){2}
          {\line(1,0){\value{temp}}}
       \setcounter{delta}{3*\value{tapeboxhight}}
       \setcounter{temp}{2*\value{tapeboxhight}}
       \multiput(\value{curx},\value{cury})(\value{delta},0){2}
          {\line(0,1){\value{temp}}}
       \setcounter{cury}{5*\value{tapeboxhight}}
       \multiput(\value{curx},\value{cury})(\value{delta},0){2}
          {\line(0,1){\value{temp}}}
       \setcounter{cury}{4*\value{tapeboxhight}+\value{tapeboxhight}/2}
       \setcounter{fstwidth}{2*\value{tapeboxhight}}
       \multiput(\value{curx},\value{cury})(\value{delta},0){2}
          {\oval(\value{fstwidth},\value{tapeboxhight})}
       \put(\value{curx},\value{cury}){\makebox(0,0){\em FST$_1$}}%
       \setcounter{curx}{4*\value{tapeboxhight}}
       \put(\value{curx},\value{cury}){\makebox(0,0){\em FST$_2$}}%
       \setcounter{curx}{9*\value{tapeboxhight}}
       \setcounter{cury}{2*\value{tapeboxhight}}
       \setcounter{delta}{3*\value{tapeboxhight}}
       \multiput(\value{curx},\value{cury})(0,\value{delta}){2}
          {\line(0,1){\value{temp}}}
       \setcounter{cury}{4*\value{tapeboxhight}+\value{tapeboxhight}/2}
       \put(\value{curx},\value{cury})
          {\oval(\value{fstwidth},\value{tapeboxhight})}
       \put(\value{curx},\value{cury}){\makebox(0,0){\em FST$_n$}}%
       \setcounter{curx}{6*\value{tapeboxhight}+\value{tapeboxhight}/2}
       \put(\value{curx},\value{cury}){\makebox(0,0){$\cdots$}}%
    \end{picture}%
   }
\newcommand{\paralleltransducerstwo}%
   {\begin{picture}(\the\value{picwidth},0)(0,-\the\value{lift})%
       \put(0,0){\framebox(\value{picwidth},\value{tapeboxhight})
             {Surface String}}%
       \setcounter{cury}{8*\value{tapeboxhight}}
       \put(0,\value{cury}){\framebox(\value{picwidth},\value{tapeboxhight})
             {Lexical String}}%
       \setcounter{curx}{\value{picwidth}/2}
       \setcounter{cury}{\value{tapeboxhight}}
       \setcounter{temp}{3*\value{tapeboxhight}}
       \setcounter{delta}{4*\value{tapeboxhight}}
       \multiput(\value{curx},\value{cury})(0,\value{delta}){2}
          {\line(0,1){\value{temp}}}
       \setcounter{cury}{4*\value{tapeboxhight}+\value{tapeboxhight}/2}
       \setcounter{fstwidth}{8*\value{tapeboxhight}}
       \put(\value{curx},\value{cury})
              {\oval(\value{fstwidth},\value{tapeboxhight})}
       \put(\value{curx},\value{cury})
              {\makebox(0,0){$FST_1 \cap FST_2 \cap \cdots \cap FST_n$}}%
    \end{picture}%
   }

\newcommand{\uniontransducers}%
   {
    \setcounter{picwidth}{10*\value{tapeboxhight}}%
    \setcounter{pichight}{9*\value{tapeboxhight}}%
    \setcounter{lift}{\value{pichight}/-2}%
    \rule[\the\value{lift}pt]{0 pt}{\the\value{pichight}pt}%
    \begin{picture}(\the\value{picwidth},0)(0,-\the\value{lift})%
       \put(0,0){\framebox(\value{picwidth},\value{tapeboxhight})
             {Surface String}}%
       \setcounter{curx}{2*\value{tapeboxhight}+\value{tapeboxhight}/2}
       \setcounter{delta}{2*\value{tapeboxhight}+\value{tapeboxhight}/2}
       \multiput(\value{curx},0)(\value{delta},0){3}
          {\dashbox{.75}(0,\value{tapeboxhight}){}}
       \setcounter{cury}{8*\value{tapeboxhight}}
       \put(0,\value{cury}){\framebox(\value{picwidth},\value{tapeboxhight})
             {Lexical String}}%
       \setcounter{cury}{8*\value{tapeboxhight}}
       \multiput(\value{curx},\value{cury})(\value{delta},0){3}
          {\dashbox{.75}(0,\value{tapeboxhight}){}}
       \setcounter{curx}{\value{tapeboxhight}}
       \setcounter{cury}{\value{tapeboxhight}}
       \setcounter{delta}{3*\value{tapeboxhight}}
       \multiput(\value{curx},\value{cury})(\value{delta},0){2}
          {\line(0,1){\value{delta}}}
       \setcounter{cury}{5*\value{tapeboxhight}}
       \multiput(\value{curx},\value{cury})(\value{delta},0){2}
          {\line(0,1){\value{delta}}}
       \setcounter{cury}{4*\value{tapeboxhight}+\value{tapeboxhight}/2}
       \setcounter{fstwidth}{2*\value{tapeboxhight}}
       \multiput(\value{curx},\value{cury})(\value{delta},0){2}
          {\oval(\value{fstwidth},\value{tapeboxhight})}
       \put(\value{curx},\value{cury}){\makebox(0,0){\em FST$_1$}}%
       \setcounter{curx}{4*\value{tapeboxhight}}
       \put(\value{curx},\value{cury}){\makebox(0,0){\em FST$_2$}}%
       \setcounter{curx}{9*\value{tapeboxhight}}
       \setcounter{cury}{\value{tapeboxhight}}
       \setcounter{delta}{4*\value{tapeboxhight}}
       \multiput(\value{curx},\value{cury})(0,\value{delta}){2}
          {\line(0,1){\value{temp}}}
       \setcounter{cury}{4*\value{tapeboxhight}+\value{tapeboxhight}/2}
       \put(\value{curx},\value{cury})
          {\oval(\value{fstwidth},\value{tapeboxhight})}
       \put(\value{curx},\value{cury}){\makebox(0,0){\em FST$_n$}}%
       \setcounter{curx}{6*\value{tapeboxhight}+\value{tapeboxhight}/2}
       \put(\value{curx},\value{cury}){\makebox(0,0){$\cdots$}}%
    \end{picture}%
    ~$\Longrightarrow$~%
    \begin{picture}(\the\value{picwidth},0)(0,-\the\value{lift})%
       \put(0,0){\framebox(\value{picwidth},\value{tapeboxhight})
             {Surface String}}%
       \setcounter{curx}{2*\value{tapeboxhight}+\value{tapeboxhight}/2}
       \setcounter{delta}{2*\value{tapeboxhight}+\value{tapeboxhight}/2}
       \multiput(\value{curx},0)(\value{delta},0){3}
          {\dashbox{.75}(0,\value{tapeboxhight}){}}
       \setcounter{cury}{8*\value{tapeboxhight}}
       \put(0,\value{cury}){\framebox(\value{picwidth},\value{tapeboxhight})
             {Lexical String}}%
       \setcounter{cury}{8*\value{tapeboxhight}}
       \multiput(\value{curx},\value{cury})(\value{delta},0){3}
          {\dashbox{.75}(0,\value{tapeboxhight}){}}
       \setcounter{curx}{\value{picwidth}/2}
       \setcounter{cury}{4*\value{tapeboxhight}+\value{tapeboxhight}/2}
       \setcounter{fstwidth}{8*\value{tapeboxhight}}
       \put(\value{curx},\value{cury})
              {\oval(\value{fstwidth},\value{tapeboxhight})}
       \put(\value{curx},\value{cury})
              {\makebox(0,0){$FST_1 \cup FST_2 \cup \cdots \cup FST_n$}}%
       \setcounter{cury}{5*\value{tapeboxhight}}
       \setcounter{temp}{4*\value{tapeboxhight}}
       \put(\value{curx},\value{cury}){\line(-4,3){\value{temp}}}
       \put(\value{curx},\value{cury}){\line(4,3){\value{temp}}}
       \put(\value{curx},\value{cury}){\line(-1,3){\value{tapeboxhight}}}
       \put(\value{curx},\value{cury}){\line(1,3){\value{tapeboxhight}}}
       \setcounter{cury}{4*\value{tapeboxhight}}
       \setcounter{temp}{4*\value{tapeboxhight}}
       \put(\value{curx},\value{cury}){\line(-4,-3){\value{temp}}}
       \put(\value{curx},\value{cury}){\line(4,-3){\value{temp}}}
       \put(\value{curx},\value{cury}){\line(-1,-3){\value{tapeboxhight}}}
       \put(\value{curx},\value{cury}){\line(1,-3){\value{tapeboxhight}}}
    \end{picture}%
   }

\newcommand{\katajakoskenniemi}%
   {
    \setcounter{picwidth}{10*\value{tapeboxhight}}%
    \setcounter{pichight}{9*\value{tapeboxhight}}%
    \setcounter{lift}{\value{pichight}/-2}%
    \rule[\the\value{lift}pt]{0 pt}{\the\value{pichight}pt}%
    \begin{picture}(\the\value{picwidth},0)(0,-\the\value{lift})%
       \put(0,0){\makebox(\value{picwidth},\value{tapeboxhight})
             {Surface Representation}}%
       \setcounter{cury}{4*\value{tapeboxhight}}
       \setcounter{temp}{8*\value{tapeboxhight}}
       \put(\value{tapeboxhight},\value{cury})
             {\makebox(\value{temp},\value{tapeboxhight})
             {Lexical Representation}}%
       \setcounter{cury}{8*\value{tapeboxhight}}
       \put(0,\value{cury}){\makebox(\value{picwidth},\value{tapeboxhight})
             {Lexical Entries (Morphemes)}}%
       \setcounter{curx}{\value{picwidth}/2}
       \setcounter{cury}{\value{tapeboxhight}}
       \setcounter{delta}{2*\value{tapeboxhight}}
       \multiput(\value{curx},\value{cury})(0,\value{delta}){4}
          {\line(0,1){\value{tapeboxhight}}}
       \setcounter{cury}{2*\value{tapeboxhight}+\value{tapeboxhight}/2}
       \setcounter{delta}{4*\value{tapeboxhight}}
       \setcounter{fstwidth}{10*\value{tapeboxhight}}
       \multiput(\value{curx},\value{cury})(0,\value{delta}){2}
          {\oval(\value{fstwidth},\value{tapeboxhight})}
       \put(\value{curx},\value{cury}){\makebox(0,0){\sc Two-Level Rules}}%
       \setcounter{cury}{6*\value{tapeboxhight}+\value{tapeboxhight}/2}
       \put(\value{curx},\value{cury}){\makebox(0,0){\sc Lexicon Component}}%
    \end{picture}%
   }

\newcommand{\environbar}{\underline{\hspace*{1.5em}}\ }

\newcommand{\phonrule}[4]%
   {#1 {}$\rightarrow${} #2 / #3 \environbar #4}

\newcommand{\tlr}[7]%
   {\begin{tabular}{cccccc}%
      {}#5&--&#6&--&#7&#4 \\
      {}#1&--&#2&--&#3&
   \end{tabular}%
   \vspace{.1in}}

\newcommand{\tlrf}[8]%
   {\begin{tabular}{cccccc}%
      {}#5&--&#6&--&#7&#4\\
      {}#1&--&#2&--&#3\\
      \multicolumn{6}{l}{{\sf Features:} {\tt #8}}
   \end{tabular}
   \vspace{.1in}}

\newcommand{\tlrc}[8]%
   {\begin{tabular}{cccccc}%
      {}#5&--&#6&--&#7&#4 \\
      {}#1&--&#2&--&#3& \\
      \multicolumn{6}{l}{{\sf where} #8}%
   \end{tabular}%
   \vspace{.1in}}

\newcommand{\tlrt}[8]%
   {\tlrc#1#2#3#4#5#6#7#8}

\newcommand{\tlrule}[9]
   {\begin{tabbing}%
       tl\_rule({\tt #1},
                  \= {\tt #2}, \= {\tt #3}, \= {\tt #4}, {\tt #5},\\%
                  \> {\tt #6}, \> {\tt #7}, \> {\tt #8},\\%
                  \> \restoftlrule#9END%
    \end{tabbing}%
   }

\newcommand{\tlruleintab}[9]
   {tl\_rule({\tt #1},
                  \= {\tt #2}, \= {\tt #3}, \= {\tt #4}, {\tt #5},\\%
             \>   \> {\tt #6}, \> {\tt #7}, \> {\tt #8},\\%
             \>   \> \restoftlrule#9END%
   }

\newcommand{\tlruleintablong}[9]
   {tl\_rule({\tt #1},
                  \= {\tt #2}, {\tt #3}, {\tt #4}, {\tt #5},
                     {\tt #6}, {\tt #7}, {\tt #8},\\%
              \>  \> \restoftlrule#9END%
   }

\def\restoftlrule#1|#2END%
  {{\tt #1}, {\tt #2}).}

%

\newcounter{examplectr}
\newcounter{subexamplectr}
%
\newenvironment{ex}%
   {\addtocounter{examplectr}{1}
     \setcounter{subexamplectr}{0}
\vspace{.1in}     \begin{list}
       {(\arabic{examplectr})}%
       {\setlength{\topsep}{0in}
	\setlength{\leftmargin}{0.45in}
	\setlength{\labelsep}{0.075in}}
       \item
   }%
   {
    \end{list}\vspace{.1in}}
%
\newenvironment{subex}%
   { \addtocounter{subexamplectr}{1}
     \begin{list}
       {\alph{subexamplectr}.}%
       {\setlength{\topsep}{-\parskip}
	\setlength{\leftmargin}{0.175in}
	\setlength{\labelsep}{0.075in}}
       \item
   }%
   {\end{list}}
%
\newcommand{\exnum}[2]{\addtocounter{examplectr}{#1}(\arabic{examplectr}{#2})\addtocounter{examplectr}{-#1}}

\author{George Anton Kiraz\thanks{Supported by a Benefactor Studentship
from St John's College. This research was done under the supervision of
Dr Stephen G. Pulman.}\\
University of Cambridge (St John's College) \\
Computer Laboratory \\
Pembroke Street \\ 
Cambridge CB2 1TP \\
{\tt George.Kiraz@cl.cam.ac.uk}}
\title{Computing Prosodic Morphology}

\maketitle

\begin{abstract}
This paper establishes a framework under which various aspects of 
prosodic morphology, such as templatic morphology and infixation, can be 
handled under two-level theory using an implemented multi-tape two-level model.
The paper provides a new computational analysis of root-and-pattern 
morphology based on prosody.
\end{abstract}

\section{Introduction}
\label{intro}

Prosodic Morphology \cite[et seq.]{McCarthyPrince:86} provides
adequate means for describing non-linear phenomena such as infixation,
reduplication and templatic morphology. Standard two-level systems
proved to be cumbersome in describing such operations -- see
\cite[p.~159 ff.]{Sproat:92} for a discussion.  Multi-tape two-level
morphology \cite[et.~seq.]{Kay:87,Kiraz:94Coling} addresses various
issues in the domain of non-linear morphology: It has been used in
analysing root-and-pattern morphology \cite{Kiraz:94Coling}, the
Arabic broken plural phenomenon \cite{Kiraz:96ICEMCO}, and error
detection in non-concatenative strings \cite{Bowden:95}.  The purpose
of this paper is to demonstrate how non-linear operations which are
motivated by prosody can also be described within this framework,
drawing examples from Arabic.

The analysis of Arabic presented here differs from earlier
computational accounts in that it employs new linguistic descriptions
of Arabic morphology, viz.~moraic and affixational theories
\cite{McCarthy:90b,McCarthy:93}.  The former argues that a different
vocabulary is needed to represent the pattern morpheme according to
the Prosodic Morphology Hypothesis (see \S\ref{pm}), contrary to the
earlier CV model where templates are represented as sequences of Cs
(consonants) and Vs (vowels).  The latter departed radically from the
notion of root-and-pattern morphology in the description of the Arabic
verbal stem (see \S\ref{infix}). 

The choice of the linguistic model depends on the application in
question and is left for the grammarian. The purpose here is to
demonstrate that multi-tape two-level morphology is adequate for
representing these various linguistic models.

The following convention has been adopted. Morphemes are represented
in braces, \{ \}, and surface forms in solidi, / /. In listings of
grammars and lexica, variables begin with a capital letter.

The structure of the paper is as follows: Section~\ref{templatic}
demonstrates how Arabic templatic morphology can be analysed by
prosodic terms, and section~\ref{infix} looks into infixation;
finally, section~\ref{conc} provides some concluding remarks. The rest
of this section introduces prosodic morphology and establishes the
computational framework behind this presentation.

\subsection{Prosodic Morphology}
\label{pm}

There are three essential principles in prosodic morphology
\cite{McCarthy:90a,McCarthy:93-pm}. They are:
\begin{ex}
   \begin{subex} 
      {\sc Prosodic Morphology Hypothesis}. \ Templates are
      defined in terms of the authentic units of prosody: mora (\Mor),
      syllable (\Syl), foot (Ft), prosodic word (PrWd).
   \end{subex}
   \begin{subex}
      {\sc Template Satisfaction Condi-\break tion}. \ Satisfaction of templates
      constraints is obligatory and is determined by the principles of
      prosody, both universal and language-specific.
   \end{subex}
   \begin{subex}
      {\sc Prosodic Circumscription}. \ The domain to which morphological
      operations apply may be circumscribed by prosodic criteria as well
      as by the more familiar morphological ones.
   \end{subex}
\end{ex}

In the {\bf Prosodic Morphology Hypothesis}, mora is the unit of
syllabic weight; a monomoraic syllable, \Sylm, is light (L), and a
bimoraic syllable, \Sylmm, is heavy (H). The most common types of
syllables are: open light, CV, open heavy, CVV, and closed heavy,
CVC. This typology is represented graphically in \exnum{+1}{}.
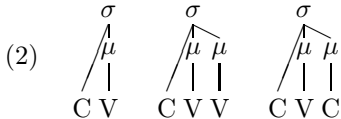
\begin{ex}
   \begin{tabular}{ccc}
       \moratree{C}{V} & \mmoratree{C}{V}{V} & \mmoratree{C}{V}{C}
   \end{tabular}
\end{ex}
Association of Cs and Vs to templates is based on the {\bf Template
Satisfaction Condition}. Association takes the following form: a node
\Syl \ always takes a C, and a mora \Mor \ takes a V; however, in
bimoraic syllables, the second \Mor \ may be associated to either a C
or a V.\footnote{Other conventions associate consonant melodies
left-to-right to the moraic nodes, followed by associating vowel
melodies to syllable-initial morae.}

{\bf Prosodic Circumscription} (PC) defines the domain of
morphological operations.  Normally, the domain of a typical
morphological operation is a grammatical category (root, stem or
word), resulting in prefixation or suffixation.  Under PC, however,
the domain of a morphological operation is a prosodically-delimited
substring within a grammatical category, often resulting in some sort
of infixation.  The essential for PC is a parsing function $\Phi$ of
the form in \exnum{+1}{}.
\begin{ex}
   {\sc Parsing Function} \\
   $\Phi$(C, E)
\end{ex}
Let B be a base (i.e. stem or word). The function $\Phi$ returns the
constituent C that sits on the edge E $\in$ \{{\tt right}, {\tt
left}\} of the base B. The result is a factoring of B into: {\bf
kernel}, designated by \kernel, which is the string returned by the
parsing function, and {\bf residue}, designated by \residue, which is
the remainder of B. The relation between \kernel\ and \residue\ is
given in \exnum{+1}{}, where \conc\ is the concatenation operator.
\begin{ex}
   {\sc Factoring of B by $\Phi$} \\
   B = \kernel\ \conc\ \residue
\end{ex}
To illustrate this, let B~=~/katab/; applying the function
$\Phi$(\Sylm, Left) on B factors it into: (i) the kernel
\kernel~=~/ka/, and (ii) the residue \residue~=~/tab/.

A morphological operation O (e.g. O~=~``Prefix \{t\}'') defined on a
base B is denoted by O(B).  There are two types of PC: {\bf positive}
(PPC) and {\bf negative} (NPC). In PPC, the domain of the operation is
the kernel \kernel; this type is denoted by \ppc\ and is defined in
\exnum{+1}{a}.  In NPC, the domain is the residue \residue; this type
is denoted by \npc\ and is defined in \exnum{+1}{b}.
\begin{ex}
   {\sc Definition of PPC and NPC}
   \begin{subex}
      PPC,  \ppc (B) = O(\kernel) \conc\ \residue
   \end{subex}
   \begin{subex}
      NPC, \npc (B) = \kernel\ \conc\ O(\residue)
   \end{subex}
\end{ex}
In other words, in PPC, O applies to the kernel \kernel, concatenating
the result with the residue \residue ; in NPC, O applies to the
residue \residue, concatenating the result with the kernel \kernel.
Examples are provided in section~\ref{infix}.

\subsection{Multi-Tape Two-Level Formalism}
\label{mttlm}

Two-level morphology \cite{Koskenniemi:83} defines two levels of
strings in recognition and synthesis: lexical strings represent
morphemes, and surface strings represent surface forms. Two-level
rules map the two strings; the rules are compiled into finite state
transducers, where lexical strings sit on one tape of the transducers
and surface strings on the other.

Multi-tape two-level morphology is an extension to standard two-level
morphology, where more than one lexical tape is allowed. The notion of
using multiple tapes first appeared in \cite{Kay:87}. Motivated by
Kay's work, \cite{Kiraz:94Coling} proposed a multi-tape two-level model. The
model adopts the formalism in \exnum{+1}{} as reported by
\cite{Pulman:93}.
\begin{ex}
      \tlr{LSC}{\sc Surf}{RSC}{\{$\Rightarrow, \Leftrightarrow$\}}
          {LLC}{\sc Lex}{RLC}\\
\end{ex}
where 
{\sc LLC} is the left lexical context,
{\sc Lex} is the lexical form,
{\sc RLC} is the right lexical context,
{\sc LSC} is the left surface context,
{\sc Surf} is the surface form,
and 
{\sc RSC} is the right surface context.

The special symbol * indicates an empty context, which is always
satisfied.  The operator $\Rightarrow$ states that {\sc Lex} {\em may}
surface as {\sc Surf} in the given context, while the operator
$\Leftrightarrow$ adds the condition that when {\sc Lex} appears in
the given context, then the surface description {\em must} satisfy
{\sc Surf}. The latter caters for obligatory rules. A lexical string
maps to a surface string iff (1) they can be partitioned into pairs of
lexical-surface subsequences, where each pair is licenced by a rule,
and (2) no partition violates an obligatory rule.

One of the extensions introduced in the multi-tape version is that all
expressions in the lexical side of the rules (i.e. {\sc LLC}, {\sc
Lex} and {\sc RLC}) are {\em n}-tuple of regular expressions of the
form (x$_1$, x$_2$, $\ldots$, x$_n$). The {\em i}th expression refers
to symbols on the {\em i}th tape.  When $n=1$, the parentheses can be
ignored; hence, $( x )$ and $x$ are equivalent.\footnote{Our
implementation interprets rules directly (see \cite{Kiraz:96ACL});
hence, we allow unequal representation of strings. If the rules were
to be compiled into automata, a genuine symbol, e.g.~0, must be
introduced by the rule compiler. For the compilation of our formalism
into automata, see \cite{Grimley-Evans:96}.}

\section{Templatic Morphology}
\label{templatic}

Templatic morphology is best exemplified in Semitic root-and-pattern
morphology. This section sets a framework under which templatic
morphology can be described using (augmented) two-level theory.  Our
presentation differs from previous proposals\footnote{Non-linear
proposals include \cite{Kay:87}, \cite{Kornai:91}, \cite{Wiebe:92},
\cite{Narayanan:93}, \cite{Bird:94} and \cite{Kiraz:94Coling}. A working
system for Arabic is reported by
\cite{Beesley:89,Beesley:90,Beesley:91}.} in that it employs prosodic
morphology in the analysis of Arabic, rather than earlier CV accounts.
Arabic verbal forms appear in \exnum{+1}{} in the passive (rare forms
are not included).
\begin{ex}
   {\sc Arabic Verbal Measures} (1-8, 10) \\
   \begin{tabular}{ll|ll}
      1 & kutib  & 6 & tukuutib\\ 
      2 & kuttib & 7 & nkutib \\
      3 & kuutib & 8 & ktutib\\
      4 & \A uktib \\
      5 & tukuttib & 10 & stuktib
   \end{tabular}
\end{ex}
\cite{McCarthy:93} points out that Arabic verbal forms are derived from
the base template in \exnum{+1}{}, which represents Measure 1.
\Sylx\ represents an extrametrical consonant;
that is, the last consonant in a stem.
\begin{ex}
   {\sc Arabic Base Template}\\
   \moratree{k}{u}\moratree{t}{i}\xmoratree{b}
\end{ex}
The remaining measures are derived from the base template by
affixation; they have no templates of their own. The simplest
operation is prefixation, e.g. \{n\} + Measure 1 $\rightarrow$
/nkutib/ (Measure 7). Measures 4 and 10 are derived in a similar
fashion, but undergo a rule of syncope as shown in \exnum{+1}{}.
\begin{ex}
   {\sc Derivation of Measures 4 and 10} \\
   Syncope: V $\longrightarrow \phi$ /[CVC \environbar CVC]$_{\mbox{stem}}$
   \begin{subex}
      Measure 4: \A u + kutib $\longrightarrow$ */\A ukutib/ 
                $\stackrel{syncope}{\longrightarrow}$ /\A uktib/
   \end{subex}
   \begin{subex}
      Measure 10: stu + kutib $\longrightarrow$ */stukutib/ 
                $\stackrel{syncope}{\longrightarrow}$ /stuktib/
   \end{subex}
\end{ex}

The following lexicon and two-level grammar demonstrate how the above
measures can be analysed under two-level theory.  The
lexicon maintains four tapes: pattern, root, vocalism and affix tapes.\\
   \begin{tabular}{lll}
   1&   \{\Sylm\Sylm\Sylx\}   & {\tt pattern:[measure=(1-8,10)]}\\
   2&   \{ktb\}              & {\tt root:[measure=(1-4,6-8,10)]}\\
   3&   \{ui\}               & {\tt vocalism:[tense=perf,}\\
    &                        & {\tt voice=pass]}\\
   4&   \{\A V\}             & {\tt verb\_affix:[measure=4]}\\
   4&   \{n\}                & {\tt verb\_affix:[measure=7]}\\
   4&   \{stV\}              & {\tt verb\_affix:[measure=10]}\\
   \end{tabular}

\noindent The first column indicates the tape on which the morpheme
sits, and the second column gives the morpheme.  Each lexical entry is
associated with a category and a feature structure of the form
{\tt cat:FS} (column 3). Feature values in parentheses are disjunctive
and are implemented using boolean vectors \cite{Mellish:88,Pulman:94}.

\{\Sylm\Sylm\Sylx\} is the base-template. \{ktb\} `notion of writing'
is the root; it may occur in all measures apart from
Measure~5.\footnote{Roots do not occur in all measures in the
literature. Each root is lexically marked with the measures it occurs
in.} \{ui\} is the perfective passive vocalism. The remaining
morphemes represent the affixes for Measures 4, 7 and 10. Notice that
the vowel in the affixes of Measures 4 and 10 is a variable V. This
makes it possible for the affix to have a different vowel according to
the mood of the following stem, e.g. [a] in /\A aktab/ (Measure 4,
active) and [u] in /\A uktib/ (Measure 4, passive).

Since the lexicon declares 4 lexical tapes, each lexical expression in
the two-level grammar must be at most a 4-tuple. A grammar for the
derivation of the cited data appears below.

\noindent {\small
\begin{tabular}{ll}
   R1&\tlr{*}{CV}{*}{$\Rightarrow$}{*}{\lab \Sylm ,C,V,\estr\rab}{*}\\
\end{tabular}
\begin{tabular}{ll}
   R2&\tlr{*}{C}{*}{$\Leftrightarrow$}{*}{\lab \Sylx,C,\estr,\estr\rab}{\lab +,+,+,\estr\rab}\\
\end{tabular}
\begin{tabular}{ll}
   R3&\tlr{*}{A}{*}{$\Rightarrow$}{*}{\lab\estr,\estr,\estr,A\rab}{*}\\
\end{tabular}
\begin{tabular}{ll}
   R4&\tlr{*}{\estr}{*}{$\Rightarrow$}{\lab X,*,\estr,\estr\rab}{\lab +,+,+,\estr\rab}{*}\\
\end{tabular}
\begin{tabular}{ll}
   R5&\tlr{*}{\estr}{*}{$\Rightarrow$}{\lab\estr,\estr,\estr,A\rab}{\lab\estr,\estr,\estr,+\rab}{*}\\
\end{tabular}
\begin{tabular}{ll}
   R6&\tlr{C$_1$V}{C}{C$_2$V$_1$C$_3$}{$\Leftrightarrow$}
                {*}{\lab \Sylm ,C,V,\estr\rab}{*} \\
\end{tabular}
}
\noindent where C$_i$={\tt radical}, V$_i$={\tt vowel}, A={\tt verbal affix},
and X~$\neq$~+.

Rule R1 handles monomoraic syllables mapping (\Sylm ,C,V,\estr) on the
lexical tapes to CV on the surface tape. Rule R2 maps the
extrametrical consonant in a stem (i.e. the last consonant in a stem)
to the surface. Rule R3 maps an affix symbol from the fourth tape to
the surface. Rules R4 and R5 delete the boundary symbols from stems
and affixes, respectively. Finally, rule R6 simulates the syncope rule
in \exnum{0}{}; note that V in LSC must unify with V in {\sc Lex},
ensuring that the vowel of the affix has the same quality as that of
the stem, e.g. /\A aktab/ and /\A u+ktib/ (Measure~4).

The two-level analysis of the cited forms appears below -- ST
= surface tape, PT = pattern tape, RT = root tape, VT = vocalism tape,
and AT = affix tape.

   \begin{tabular}{ll} 
      Measure 1 & Measure 4 \\
      \mttlm{{ku}{ti}b{}:ST}{1124}
         {{\Sylm}{\Sylm}{\Sylx}+:PT|ktb+:RT|ui-+:VT|----:AT} &
      \mttlm{{\A}u{}k{ti}b{}:{}}{3356124}
         {---{\Sylm}{\Sylm}{\Sylx}+:{}|---ktb+:{}|---ui-+:{}|{\A}u+----:{}} \\
& \\
      Measure 7 & Measure 10 \\ 
     \mttlm{n{}{ku}{ti}b{}:ST}{351124}
         {--{\Sylm}{\Sylm}{\Sylx}+:PT|--ktb+:RT|--ui-+:VT|n+----:AT} &
      \mttlm{stu{}k{ti}b{}:{}}{33356124}
         {----{\Sylm}{\Sylm}{\Sylx}+:{}|----ktb+:{}|----ui-+:{}|stu+----:{}}
   \end{tabular} \mttlmsetwidth{12}%
\smallskip

\noindent The numbers between the two levels indicate the rule numbers in
\exnum{-1}{} which sanction the sequences. The remaining Measures
involve infixation and are discussed in the next section.

\section{Infixation}
\label{infix}

Standard two-levels models can describe some classes of infixation,
but resorting to the use of {\em ad hoc} diacritics which have no
linguistic significance, e.g.~\cite[p.~156]{Antworth:90}.  This
section presents a framework for describing infixation rules using our
multi-tape two-level formalism.  This is illustrated here by analysing
Measures 2 and 8 of the Arabic verb. Measure 2, /kuttib/, is derived
by prefixing a mora to the base template under NPC.  The operation is
O~=~`prefix \Mor' and the rule is \npc(\Sylm, Left). The new mora is
filled by the spreading of the adjacent (second) consonant. The
steps of the derivation are:
   \begin{tabbing}
      \npc ({\tt kutib}) \= = {\tt kutib}:$\Phi$ * O({\tt kutib}:$\Phi$)
                         \ \ \ \ \ \\ 
      \> = {\tt ku} * O({\tt tib}) \\ 
      \> = {\tt ku} * {\tt \Mor tib}\\
      \> = {\tt ku} * {\tt ttib}    \\
      \> = {\tt kuttib}
   \end{tabbing}

Measure 8, /ktutib/, is derived by the affixation of a \{t\} to the
base template under NPC. The operation is O~=~`prefix \{t\}'; the rule
is \npc(C, Left), where C is a consonant. The process is:
   \begin{tabbing}
      \npc ({\tt kutib}) \= = {\tt kutib}:$\Phi$ * O({\tt kutib}:$\Phi$)
                         \\
      \> = {\tt k} * O({\tt utib}) \\
      \> = {\tt k} * {\tt tutib} \\
      \> = {\tt ktutib}
   \end{tabbing}

The following two-level grammar builds on the one discussed in
section~\ref{templatic}. The following lexical entry gives the
Measure 8 morphemes.

\noindent \begin{tabular}{lll}
    4 & \{t\}     & {\tt verb\_affix:[measure=8]}
   \end{tabular}
\smallskip

The additional two-level rules are:

\noindent {\small \begin{tabular}{llll}
R7 & \tlrf{*}{C}{*}{$\Rightarrow$}
                 {\lab \Sylm,C$_1$,V$_1$,\estr\rab}{\estr}{\lab \Sylm,C,*,\estr\rab}{[measure=(2,5)]}\\
R8 & \tlrf{*}{CAV}{*}{$\Rightarrow$}{*}{\lab \Sylm,C,V,A\rab}{*}
                 {[measure=8]}
   \end{tabular}
}
\noindent where C$_i$={\tt radical}, V$_i$={\tt vowel}, A={\tt verbal affix},
and X~$\neq$~+.

Rules R7-R8 are measure-specific.  Each rule is associated with a
feature structure which must unify with the feature structures of the
affected lexical entries.  This ensures that each rule is applied
only to the proper measure.  

R7 handles Measure 2; it represents the operation O~=~`prefix \Mor '
and the rule \npc(\Sylm, Left) by placing B:$\Phi$ in LLC and the
residue B/$\Phi$ in RLC, and inserting a consonant C (representing
\Mor ) on the surface. The filling of \Mor \ by the spreading of the
second radical is achieved by the unification of C in {\sc Lex} with
C in RLC.

R8 takes care of Measure 8; it represents the operation O~=~`prefix
\{t\}' and the rule \npc(C, Left). Note that one cannot place \kernel\
and \residue\ in LLC and RLC, respectively, as the case in R7 because
the parsing function cuts into the first syllable.

One remaining Measure has not been discussed, Measure 3.  It is
derived by prefixing the base template with \Mor .
The process is as follows:

\noindent\begin{tabular}{lll}
 \moratree{k}{u}\moratree{t}{i}\xmoratree{b} & $\longrightarrow$ &
   \Mor \ + \moratree{k}{u}\moratree{t}{i}\xmoratree{b}\\
&\\
  & $\longrightarrow$ & \mmoratree{k}{u}{u}\moratree{t}{i}\xmoratree{b}
\end{tabular}
\smallskip

The corresponding two-level rule follows. It adds a
\Mor \ by lengthening the vowel V into VV. 

\noindent {\small \begin{tabular}{llll}
 R9 & \tlrf{*}{CVV}{*}{$\Rightarrow$}
                 {*}{\lab \Sylm,C,V,\estr\rab}{*}{[measure=(3,6)]}
\end{tabular}
}

The two-level derivations are:

\noindent\mttlmsetwidth{17}
    \begin{tabular}{ll}
       Measure 2 &        \mttlm{{ku}t{ti}b{}:ST}{17124}
             {{\Sylm}-{\Sylm}{\Sylx}+:PT|k-tb+:RT|u-i-+:VT|-----:AT}\\
& \\
      Measure 3 & \mttlm{{kuu}{ti}b{}:ST}{9124}
             {{\Sylm}{\Sylm}{\Sylx}+:PT|ktb+:RT|ui-+:VT|----:AT}\\
& \\
     Measure 8 & \mttlm{{ktu}{}{ti}b{}:ST}{85124}
             {{\Sylm}-{\Sylm}{\Sylx}+:PT|k-tb+:RT|u-i-+:VT|t+---:AT}
    \end{tabular}
    \mttlmsetwidth{12}%
\smallskip

Finally, Measures 5 and 6 are derived by prefixing \{tu\} to Measures
 2 and 3, respectively.

 \section{Conclusion}
 \label{conc}

 This paper have demonstrated that multi-tape two-level systems offer
 a richer and more powerful devices than those in standard two-level
 models. This makes the multi-tape version capable of modelling
 non-linear operations such as infixation and templatic morphology.

 The rules and lexica samples reproduced here are based on a larger 
 morphological grammar written for the {\tt SemHe} implementation
 (a multi-tape two-level system) -- for a full description of the system,
 see \cite{Kiraz:96ACL,Kiraz:thesis}.


\end{document}